\documentclass[sigconf,edbt]{acmart-edbt2024}

\def\BibTeX{{\rm B\kern-.05em{\sc i\kern-.025em b}\kern-.08em
    T\kern-.1667em\lower.7ex\hbox{E}\kern-.125emX}}

\usepackage{booktabs} % For formal tables

\setcopyright{rightsretained}

\acmDOI{}

\acmISBN{978-3-89318-095-0}

\acmConference[EDBT 2023]{27th International Conference on Extending Database Technology (EDBT)}{25th March-28th March, 2024}{Paestum, Italy}
\acmYear{2024}
\theoremstyle{plain}
\newtheorem{thm}{Theorem}% reset theorem numbering for each chapter
\theoremstyle{definition}
\newtheorem{defn}[thm]{Definition} % definition numbers are dependent on theorem numbers

\usepackage[linesnumbered,ruled,vlined]{algorithm2e}
\usepackage{enumitem}
\usepackage{multirow}
\usepackage{multicol}
\usepackage{makecell}
\usepackage{graphicx}
\usepackage[normalem]{ulem}
\usepackage[font=footnotesize]{subfig}
\usepackage{amsthm}
\usepackage{amsmath}

\begin{document}
%%
%% The "title" command has an optional parameter,
%% allowing the author to define a "short title" to be used in page headers.
%\title{\textsf{MAGNETO}: Edge AI for Human Activity Recognition with New Activities Integration On the Fly}
\title{\textsf{MAGNETO}: Edge AI for Human Activity Recognition - Privacy and Personalization}

\author{Jingwei Zuo}
\affiliation{Technology Innovation Institute\\
  \city{Abu Dhabi}
  \country{UAE}}
\email{jingwei.zuo@tii.ae}

\author{George Arvanitakis}
\affiliation{Technology Innovation Institute\\
  \city{Abu Dhabi}
  \country{UAE}}
\email{george.arvanitakis@tii.ae}

\author{Mthandazo Ndhlovu}
\affiliation{%
  \institution{Technology Innovation Institute}
  \city{Abu Dhabi}
  \country{UAE}}
\email{mthandazo.ndhlovu@tii.ae}

\author{Hakim Hacid}
\affiliation{Technology Innovation Institute\\
  \city{Abu Dhabi}
  \country{UAE}}
\email{hakim.hacid@tii.ae}

%%
%% The abstract is a short summary of the work to be presented in the
%% article.
\begin{abstract}
%JZ: To show the context before presenting MAGNETO - why it is needed by the community?
%Edge Artificial Intelligence (Edge AI)
Human activity recognition (HAR) is a well-established field, significantly advanced by modern machine learning (ML) techniques. While companies have successfully integrated HAR into consumer products, they typically rely on a predefined activity set, which limits personalizations at the user level (edge devices). Despite advancements in Incremental Learning for updating models with new data, this often occurs on the Cloud, necessitating regular data transfers between cloud and edge devices, thus leading to data privacy issues. 
%\textit{sMArt sensinG for humaN activity rEcogniTiOn},
In this paper, we propose \textsf{MAGNETO}, an Edge AI platform that pushes HAR tasks from the Cloud to the Edge. \textsf{MAGNETO} allows incremental human activity learning directly on the Edge devices, without any data exchange with the Cloud. This enables strong privacy guarantees, low processing latency, and a high degree of personalization for users. In particular, we demonstrate \textsf{MAGNETO} in an Android device, validating the whole pipeline from data collection to result visualization.
%is equipped with the ability to incrementally learn new situations, i.e., activities/classes, by capturing a few extra user data in order to (i) re-calibrate an activity to better match user's personal style or (ii) re-train the model to learn new custom activities according to user’s habits. \textsf{MAGNETO} operates both the re-calibration and the re-training processes on the edge device without any data exchange with the cloud, ensuring a full privacy, a very low latency, and a high degree of flexibility. 
%Last but not least, the amount of samples that the user needs to capture in order for \textsf{MAGNETO} to learn a new activity is just the order of a few seconds of recording.

%JZ: To show technical contributions or briefly mention adopted techniques/models
%JZ: To show how the audience will experience/operate during the demo.
\end{abstract}

%%
%% This command processes the author and affiliation and title
%% information and builds the first part of the formatted document.
\maketitle
\section{Introduction}
Human Activity Recognition (HAR) tasks has been largely investigated in the past decades. Numerous researches have studied the HAR problem from different aspects, from data collection, learning models, to post-processing and result interpretation~\cite{ANTAR2021146}. 

\begin{figure}[ht]
    \centering
    \includegraphics[width=1\linewidth]{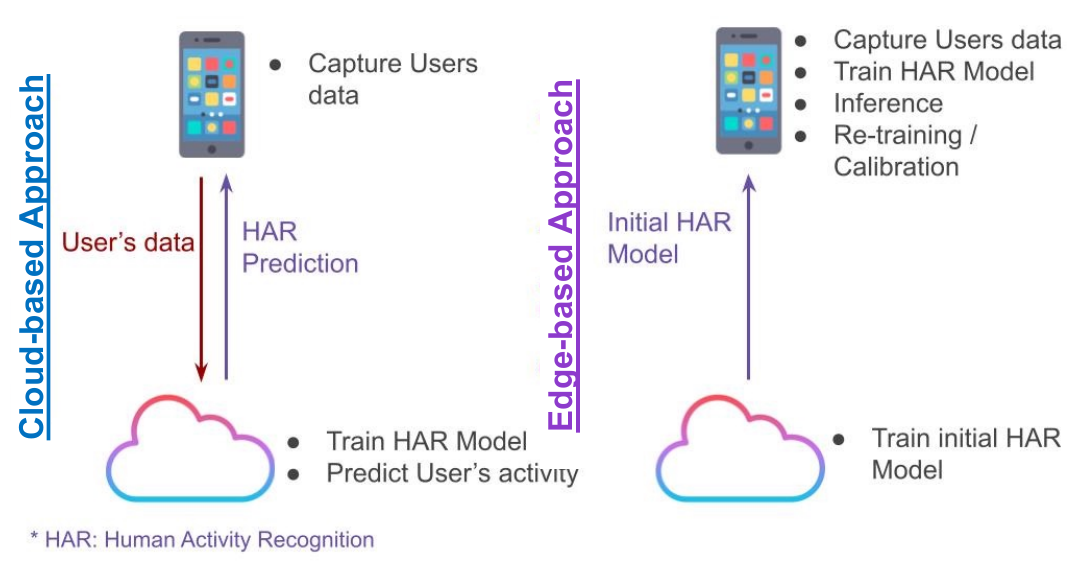}
    \caption{left) Human Activity Recognition (HAR) protocols: left) Cloud-based Approach, constant communication between Cloud and Edge, right) Edge-based Approach, only data transfer from Cloud to Edge is allowed, showing stronger privacy guarantees.
    }
    \label{fig:architecture_cloud_edge_HAR}
    \vspace{-2em}
\end{figure}

% Shortcomings of previous research/solutions on human activity recognition
As shown in Figure~\ref{fig:architecture_cloud_edge_HAR}, traditional approaches~\cite{ANTAR2021146,googlelink, applelink} for HAR tasks are mainly Cloud-based: a classifier is trained on a predefined set of activities in a centralized Cloud environment. User's activity data collected on the Edge device is then sent to the Cloud for inference. However, this centralized, Cloud-based approach raises three main issues: (i) \textit{high latency}, due to the User-Cloud communication, (ii) \textit{lack of flexibility and personalization} to individual user’s needs, and (iii) \textit{lower privacy control}, due to the data transfer to the Cloud. 
In contrast with the conventional Cloud-based approach, Edge AI~\cite{zhou2019edge} shifts core processing tasks (e.g., model's training, inference, etc.) to the edge devices and intends to adapt AI technologies to the edge environment. In a specific use case, edge devices can collect user activity data directly and use it to update an initial model, catering well to user demands for \textit{personalization}. This approach facilitates the deployment of optimized models and services directly onto the user devices, thus ensuring rapid real-time response (\textit{low latency}) and enhanced \textit{privacy guarantees}. 

% Challenges on Edge ML
However, shifting both inference and learning process to Edge devices presents unique challenges due to the intrinsic constraints of Edge devices, including (i) Model size, which should be small enough to fit within the {Edge} but also to operate efficiently,  (ii) Data size, which should be very limited due to the low storage capabilities within the {Edge}, and (iii) Energy consumption, constraining the training process to be very efficient without excessive power consumption. Moreover, in the context of personalized HAR, since models are required to be updated continuously (i.e., \textit{personalization}), \textit{Catastrophic Forgetting}~\cite{zuo2023OnEdge} is a practical challenge when learning from the dynamic data streams. This becomes even more daunting with the limited Edge resources. 

%Therefore, our proposal, an Edge-based approach as shown in Figure \ref{fig:architecture_cloud_edge_HAR}, aims at tackling the two aforementioned issues by pushing the key processes (e.g., model (re-)training, inference, calibration, etc.) to the \textsf{Edge}, thus avoiding frequent data exchanges between the Edge and the Cloud. Therefore, it allows learning new activities directly on the Edge, based on a HAR model initially trained on the Cloud.

%JZ: this figure may be put in the Introduction show highlight the difference between cloud-based and edge-based architectures
%\begin{figure}
%    \centering
%    \includegraphics[width=0.95\linewidth]{img/ARCH_MAGNETO_2 (4).jpg}
%    \caption{(left side: Cloud-based approach) (i) activity data is captured on the \textsf{Edge} device, (ii) transferred to the Cloud, (iii) used to build (learn) a model or infer the activity, and (iv) send the conclusion back to the \textsf{Edge} device.\\
%    (right side: Our \textsf{Edge} approach) (i) an initial and basic classification model is trained on the cloud (similar to transfer learning), (ii) data is collected on the device, (iii) the \textsf{Edge} device performs the learning (adding new activities) or the inference, (iv) the device displays the conclusions.  
%    }
%    \label{fig:architecture_cloud_edge_HAR}
%\end{figure}

In this paper, considering the aforementioned challenges, we introduce \textsf{MAGNETO}, \textit{sMArt sensinG for humaN activity rEcogniTiOn}, an Edge AI platform that shows the whole pipeline of HAR tasks, covering real-time data collection, data preprocessing, model adaptation/re-training/calibration, model inference and result visualization. Importantly, all the operations are taking place on the Edge, and should not have any data exchange with
the Cloud, considering data privacy issues. Moreover, \textsf{MAGNETO} is equipped with the Incremental Learning ability for new activities, without \textit{forgetting} previously learned activities, i.e., catastrophic forgetting. To demonstrate the effectiveness of \textsf{MAGNETO}, a user-friendly Android application has been developed. This offers users an interactive and practical way to experience the platform's capabilities on a smartphone. We believe that the HAR on the Edge with incremental learning capability can enable a new area in the health care, fitness or assistant applications.

The rest of this paper is organized as follows: Section~\ref{sec:related_work} presents
the most related work to our proposal. Section~\ref{sec:systemdesign} introduces the Edge AI platform \textsf{MAGNETO}. Section~\ref{sec:demo_scenario} discusses the demonstration scenarios. We conclude and show future work in Section~\ref{sec:conclusion}. 
\section{Related work}\label{sec:related_work}
%JZ: newly added section
In this section, we show the most related work of our Edge AI platform, covering Edge Machine Learning and HAR tasks. 

\subsection{Edge Machine Learning (Edge ML)}
Edge Machine Learning (Edge ML) shows advantages on low latency and strong privacy guarantee. Edge ML can be divided into two categories: Inference and Training on the Edge. For model inference on the Edge, past studies focus on optimizing model scale and quantizing weights to reduce resource costs, employing methods like parameter pruning~\cite{han2015learning}, low-rank factorization~\cite{denton2014exploiting}, and knowledge distillation~\cite{hinton2015distilling}. Training on the Edge, which has higher resource costs, either uses tiny models or employs distributed/federated learning~\cite{yang2019federated}. The paper investigates models that can efficiently operate on Edge devices with limited resources.
The objective of \textsf{MAGNETO} is to build a complete pipeline for HAR tasks, we should note that more sophisticated techniques can be integrated into the platform incrementally.

\subsection{Human Activity Recognition (HAR)}
% competitors
%there is some work in the direction of few-shots learning~\cite{FENG2019112782, s20030825}.
Human Activity Recognition (HAR) is a well-established field. Building HAR model is basically considered as a classification task. Depending on data resources and targeted applications, the HAR model can be designed differently. One can use handcrafted features to feed any general ML models for downstream tasks, which is easy-to-deploy and requires linear processing time. More advanced work has been proposed in the Time Series Classification domain, where researchers aim to build general ML models covering various application domains~\cite{zuo2021smate}, including HAR tasks. For instance, Shapelet features~\cite{zuo2019exploring} with a kNN classifier, end-to-end models~\cite{zuo2021smate} with automatic feature extraction and selection.
We should remark that the HAR models have been integrated into various commercial service or products, such as Google platform~\cite{googlelink}, Samsung health activity trackers~\cite{samsunglink}, and Apple developer kit~\cite{applelink}. 

However, the existing work either rely on centralized learning or inference on the Cloud~\cite{zuo2021smate,zuo2019exploring}, or a pre-defined activity sets lacking support for personalization~\cite{googlelink,samsunglink,applelink}. This fall short in simultaneously addressing the crucial aspects of the Edge Learning on HAR tasks: \textit{Privacy} and \textit{Personalization}. 
%hese approaches aim to reach optimal model performance with less attention on the model's scale or complexity, which are essential in the Edge context regarding the extremely limited resources. In this paper, we consider the HAR task as an application and focus on the Edge model's incremental learning behavior. Therefore, we adopt a primary feature extractor that relies on handcrafted statistic features, requiring linear processing time. Nevertheless, more advanced feature extractors can be explored and integrated into our framework, by considering the Edge constraints. This is orthogonal to our work.

\section{Proposal: \textsf{MAGNETO}}
\label{sec:systemdesign}

\begin{figure*}
    \centering
    \includegraphics[width=0.95\linewidth]{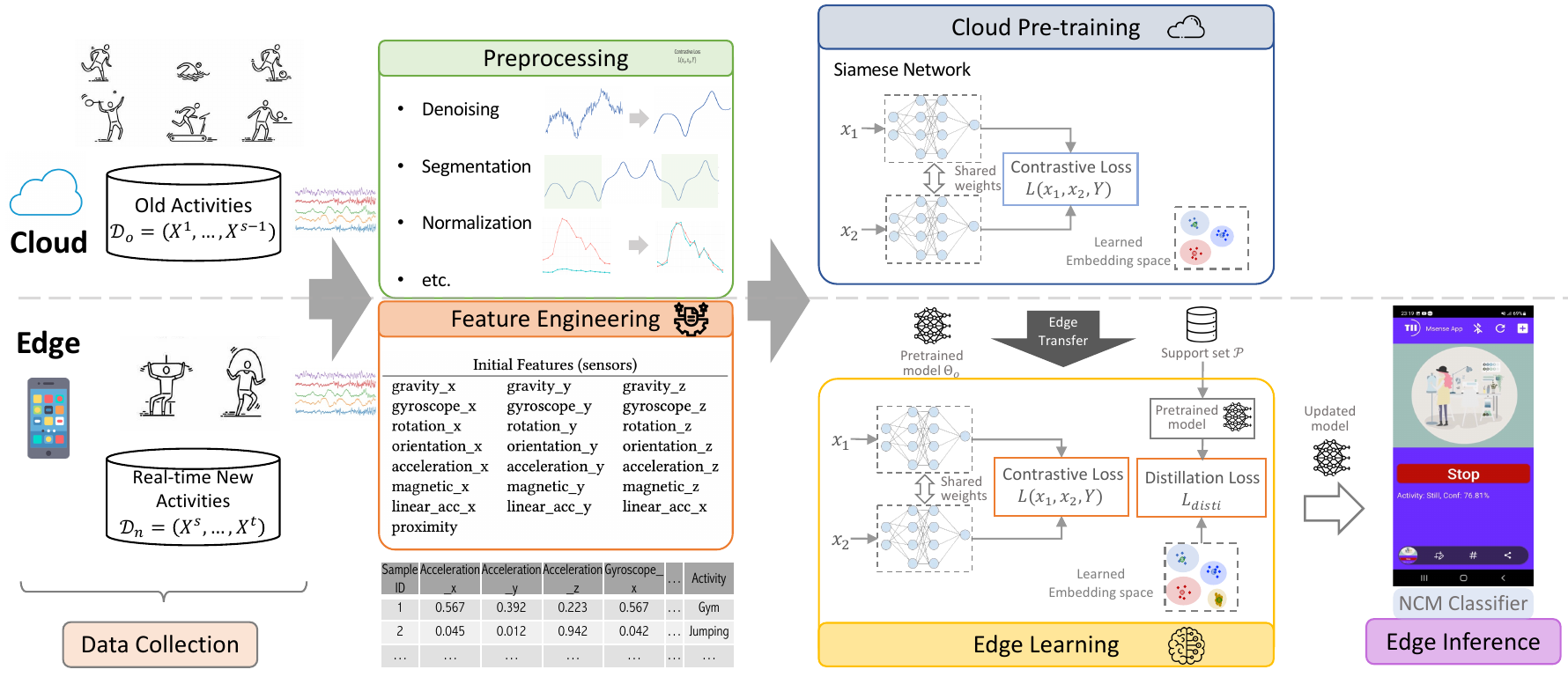}
    \caption{Global system architecture of \textsf{MAGNETO}}
    \label{fig:Magneto architecture}
\end{figure*}

%JZ: to add a global architecture of MAGNETO, Figure 1 only shows the Edge concept other than a detailed technical structure.
To answer the aforementioned challenges, we propose \textsf{MAGNETO}, \textit{sMArt sensinG for humaN activity rEcogniTiOn}, which is an Edge AI platform primarily designed for HAR tasks. Before describing its system design, we highlight two main definitions:
\begin{defn}(Privacy). Given a Cloud server and Edge device, no user data is allowed to be transferred from Edge to Cloud. However, it is less restrict to pull data from Cloud to Edge.
\end{defn}

\begin{defn}(Personalization). Given a model $\Theta_o$ trained on $\mathcal{D}_o$, user's personal data $\mathcal{D}_n = (X_{s}, ..., X_{t})$ with classes $(s, ..., t)$ will enrich $\Theta_o$ incrementally, leading to a new model $\Theta_{n}$.
\end{defn}

We should note that the \textit{Personalization} is strongly linked to the \textit{Incremental Learning} paradigm~\cite{de2021continual}. With extra user data, the model can be personalized in two ways: (i) re-calibrate an activity to be more accurate for her/his personal style or (ii) re-train the model to learn new custom activities according to user’s habits. 

\subsection{Architecture}
%Figure~\ref{fig:Magneto architecture} shows the main architecture of \textsf{MAGNETO}. Different from the Cloud-based approach with continuous data exchange between Cloud and Edge, \textsf{MAGNETO} is built primarily with an Edge-based structure
%, in which an initial HAR model is pre-trained on the Cloud as a warm starting point~\cite{ash_warm-starting_2020} for the Edge learning process. The pre-trained model benefits from the rich computation resources on the Cloud computing center, meanwhile allowing further adaptation for new-coming real-time data on the Edge.

%With limited computation resources but high flexibility, the Edge devices in MAGNETO platform provide the possibility for real-time data collection, model adaptation/re-training/calibration, and model inference. Importantly, all the operations are taking place on the Edge, and should not have any data exchange with the Cloud, considering data privacy issues.

As mentioned before, the goal of \textsf{MAGNETO} is to provide user activity estimation directly on the \textsf{Edge} device and with the capability of adding new activities in the model, without transferring any of the user’s data to the Cloud.
Figure~\ref{fig:Magneto architecture} shows the architecture and the components of \textsf{MAGNETO}, discussed below. 
%. The main components of this architecture will be explained in the rest of this section. 
To achieve its objectives, \textsf{MAGNETO} proceeds in two steps: (i) an offline step (\textit{Cloud Initialization}) and (ii) an online step (\textit{Edge Inference and Learning}). After learning a class-separable embedding space, a nearest class mean (NCM) classifier~\cite{zuo2023OnEdge} can be built to do the Edge Inference.

The \textbf{offline step} in this context is carried out on the Cloud for two main reasons: (a) resource limitations on the \textsf{Edge}, inviting us to leverage the scalable resources of the Cloud but without compromising user privacy. This task is actually similar to a transfer learning approach where a pre-trained model is used for, e.g., personalization purposes. It's essential for initializing the system, addressing the cold-start problem by providing a pre-trained model as a foundational knowledge base. This base is crucial for subsequent learning on the Edge, as it is not efficient to start from scratch.
It's important to note that in the pre-training process, no user data is transferred from Edge to Cloud. The initial model is pre-trained exclusively using open-source data that is readily available, ensuring user data privacy and security. 

During the \textbf{online step}, \textsf{MAGNETO} performs real-time inference of the user's activities based on their data, while incrementally learning from new data. This feature enhances personalization to the user's specific needs. Notably, learning from new data occurs directly on the device, eliminating the need to transmit data to the cloud. Finally, from the inference standpoint, this approach ensures minimal latency since all operations are executed directly on the \textsf{Edge}. More technical details of the online step (e.g., handling \textit{Catastrophic Forgetting issues} and \textit{Incremental Learning behaviors}) can be found in our previous papers~\cite{zuo2023OnEdge,arvanitakis2023practical}.

\subsection{Cloud Initialization}
\label{sec:edgeinference}
To empower \textsf{MAGNETO} with the best possible initial model, instead of building an initial dataset from openly available data source, we launched data collection campaigns collecting an initial sensory dataset of more than 100GB, which was stored and processed on the cloud. 
A neural network is built from the pre-processed data, targeting the prediction of existing activities, embedded in the system as an initialization step. 
%The initial data-set that has been captured from large measuring campaigns is stored on the cloud. This initial data-set, reaching a 100GB of  sensor data in our case, has been processed on the cloud. 
At the end of this step, the following items are transferred into the \textsf{Edge} device:

\begin{enumerate}[leftmargin=.2in]

\item \textbf{The pre-processing function}: We do popular pre-processing operations on raw sensor data, including denoising, segmentation, normalization, as shown in Figure~\ref{fig:Magneto architecture}. Moreover, we adopt a primary feature extractor that relies on handcrafted statistic features~\cite{zuo2023OnEdge}, requiring linear processing time. Nevertheless, more advanced feature extractors can be explored and integrated into our framework, by considering the Edge constraints. This is orthogonal to our work. 

\item \textbf{The Initial \textsf{ML} Model}: Being efficient and memory friendly, a Siamese Network-based model~\cite{Koch2015SiameseNN} with contrastive loss~\cite{NEURIPS2020_d89a66c7} is designed, which learns a class-separable embedding space. The backbone model is a simple Fully
Connected (FC) neural network with dimensions [$1024 \times 512 \times 128 \times 64 \times 128 $], which can be replaced by any other advanced networks. The lightweight model can be efficiently deployed on Edge devices for retraining, even though the training data is very limited~\cite{zuo2023OnEdge}.

\item \textbf{The support set}: As \textsf{MAGNETO} supports learning new activity data on the fly, it is necessary to keep a minimal dataset to update the learning model (mainly for handling \textit{Catastrophic Forgetting} issues~\cite{zuo2023OnEdge}). The support set, containing a limited amount of data samples which are representative for each class, can greatly reduce the resource cost for Edge devices. For instance, $200$ observations per class cost roughly $0.5 MB$ in 32-bit precision. This support set has a two-fold mission: (i) serving to calculating the class prototypes for building the $NCM$ classifier, (ii) updating the model by combining with the new activity data as training set. 
\end{enumerate}

\subsection{Edge Inference and  new activities learning }%on the \textsf{Edge}}

With the aforementioned transferred items from the cloud initialization stage, the \textsf{Edge} device is capable of performing the inference on the fly by reading its sensors and passing the captured measurements sequentially from the pre-processing function to the pre-trained model. Again, this is performed without sharing any of the user's raw information with the Cloud.

For learning new data on the Edge, either on existing or new activities, we adopt the same base model as the \textit{Cloud Initialization}, i.e., Siamese Network~\cite{Koch2015SiameseNN} with contrastive loss~\cite{NEURIPS2020_d89a66c7}. This few-shot learning approach allows updating the model with minimal data. To handle the \textit{Catastrophic Forgetting} issue~\cite{zuo2023OnEdge}, we jointly optimize the model with contrastive loss~\cite{NEURIPS2020_d89a66c7} and distillation loss \cite{hinton2015distilling}. Here are the main steps happened on \textsf{Edge} devices: 

\begin{enumerate}[leftmargin=.2in]
%TODO: to remove the exact number of recording period 
\item\textbf{Samples recording}: Users capture new activity samples that does not exist in the initial dataset, while annotating the activity, e.g., roughly $20$-$30$ seconds of recording, that will be fed into the pre-processing function.

\item \textbf{Support set update}: To keep track of the activities of users and ensure an incremental learning process, \textsf{MAGNETO} adds the freshly captured data into the support set, which serves to further re-training of the model.

\item \textbf{Model re-training}: The initial model will be updated by integrating the patterns of the newly captured data. As mentioned previously, the cost function is a combination of Contrastive and Distillation Loss, optimized on the updated support set.

\end{enumerate}

Following the re-training steps, the inference resumes as previously described. It's important to note two key aspects: first, the learning process can be repeated to accommodate the addition of multiple activities as per the user's requirements; second, calibrating an activity to more closely align with the user's behavior is a focal point of interest. This calibration mirrors the re-training process, with the distinction that the data for the targeted activity within the support set is replaced with newly acquired data.
\section{About the Demonstration}
\label{sec:demo_scenario}

\subsection{Demonstration settings}
\subsubsection{Demonstration environment}
\textsf{MAGNETO} is developed and integrated into an Android Application. The core models are implemented in PyTorch 1.6.0. The proposal will be demonstrated in an Android smartphone, freely accessible to audiences.

\subsubsection{Dataset description}
We base our demonstration on real human physical activity data collected on edge devices. 
We have launched data collection campaigns, capturing an initial dataset of more than $100GB$ of sensor data. We split the sensory data into a one-second window with roughly 120 sequential measurements from 22 mobile sensors, e.g., accelerometer, gyroscope, and magnetometer. We extract 80 statistical features. After data preprocessing, five activities with $\sim 200k$ records are collected: \textit{Drive}, \textit{E-scooter}, \textit{Run}, \textit{Still}, \textit{Walk}. The data is applied to pre-train a HAR model. On the Edge device, the real-time coming data can be processed instantly, as the preprocessing requires linear time. 

\subsection{Demonstration scenarios}
\begin{figure*}[ht]
  \begin{minipage}[t]{.175\linewidth}
    \includegraphics[width=\linewidth]{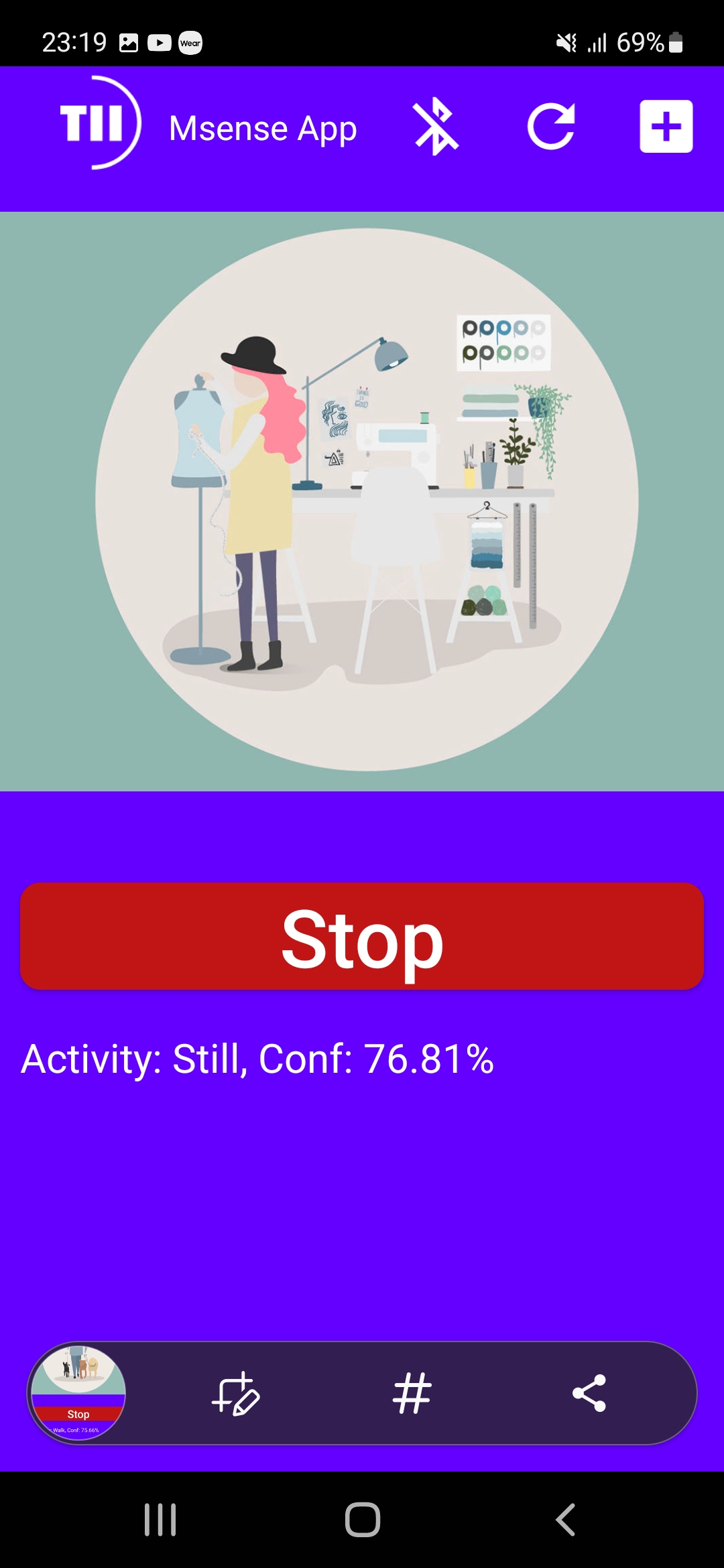}%
    \caption*{(a)}%
  \end{minipage}\hfil
  \begin{minipage}[t]{.175\linewidth}
    \includegraphics[width=\linewidth]{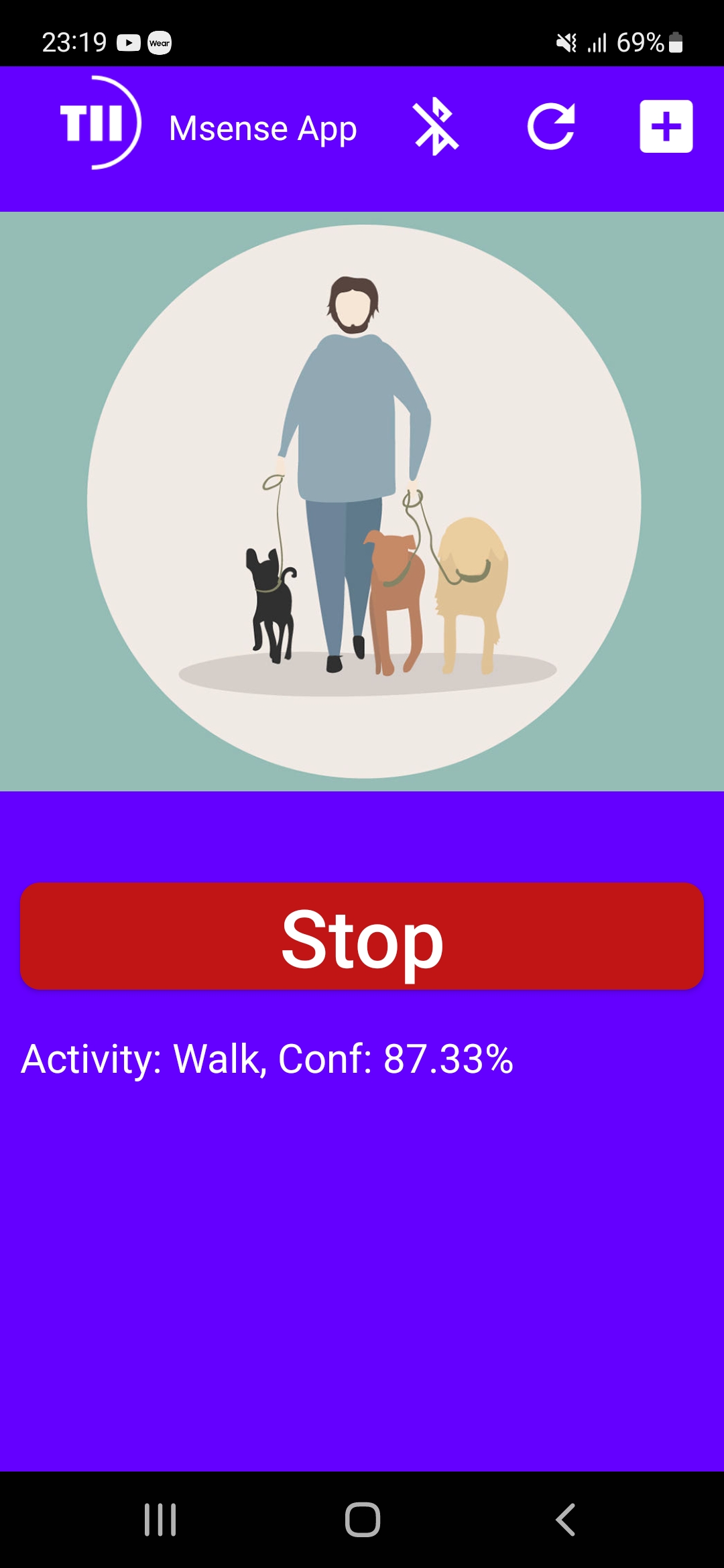}%
    \caption*{(b)}%
  \end{minipage}\hfil
    \begin{minipage}[t]{.175\linewidth}
    \includegraphics[width=\linewidth]{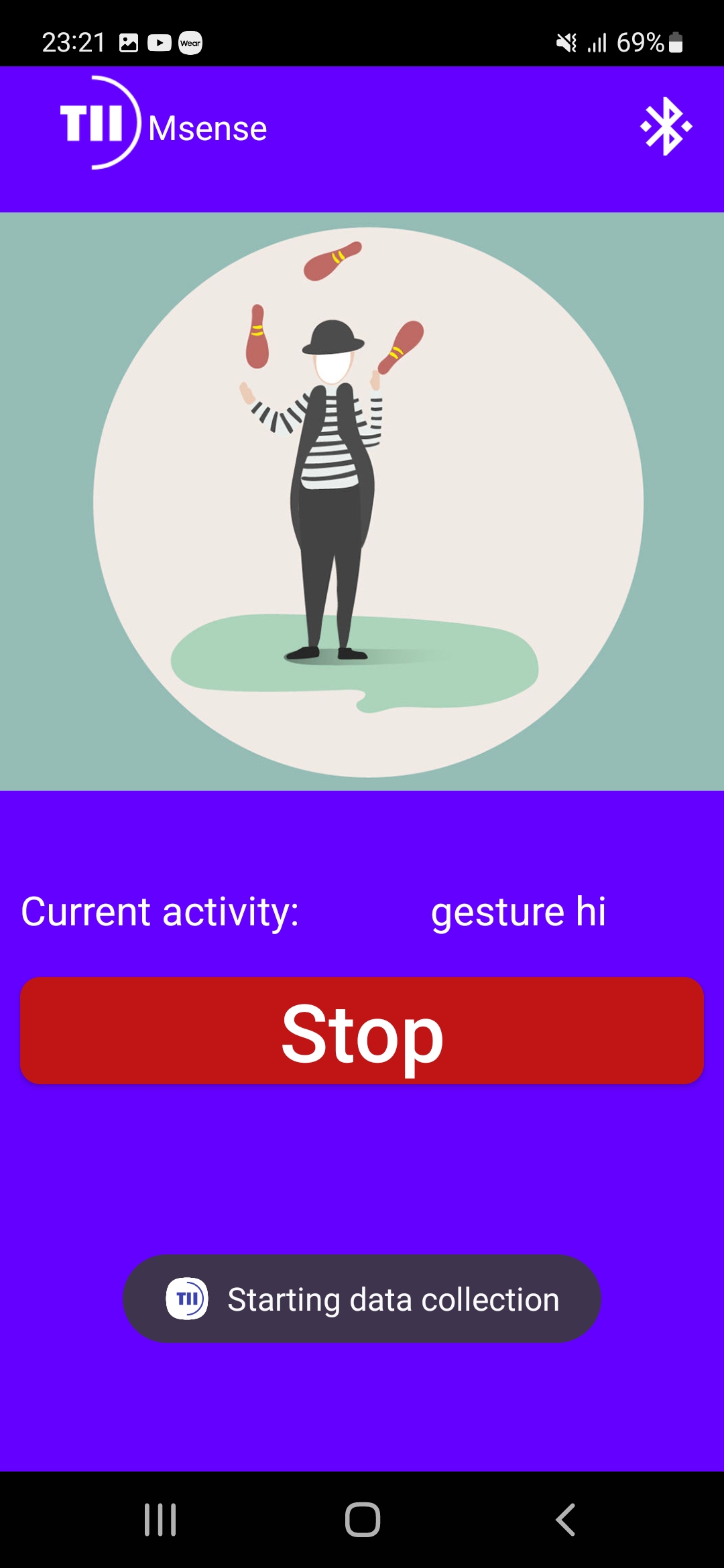}%
    \caption*{(c)}%
  \end{minipage}\hfil
    \begin{minipage}[t]{.175\linewidth}
    \includegraphics[width=\linewidth]{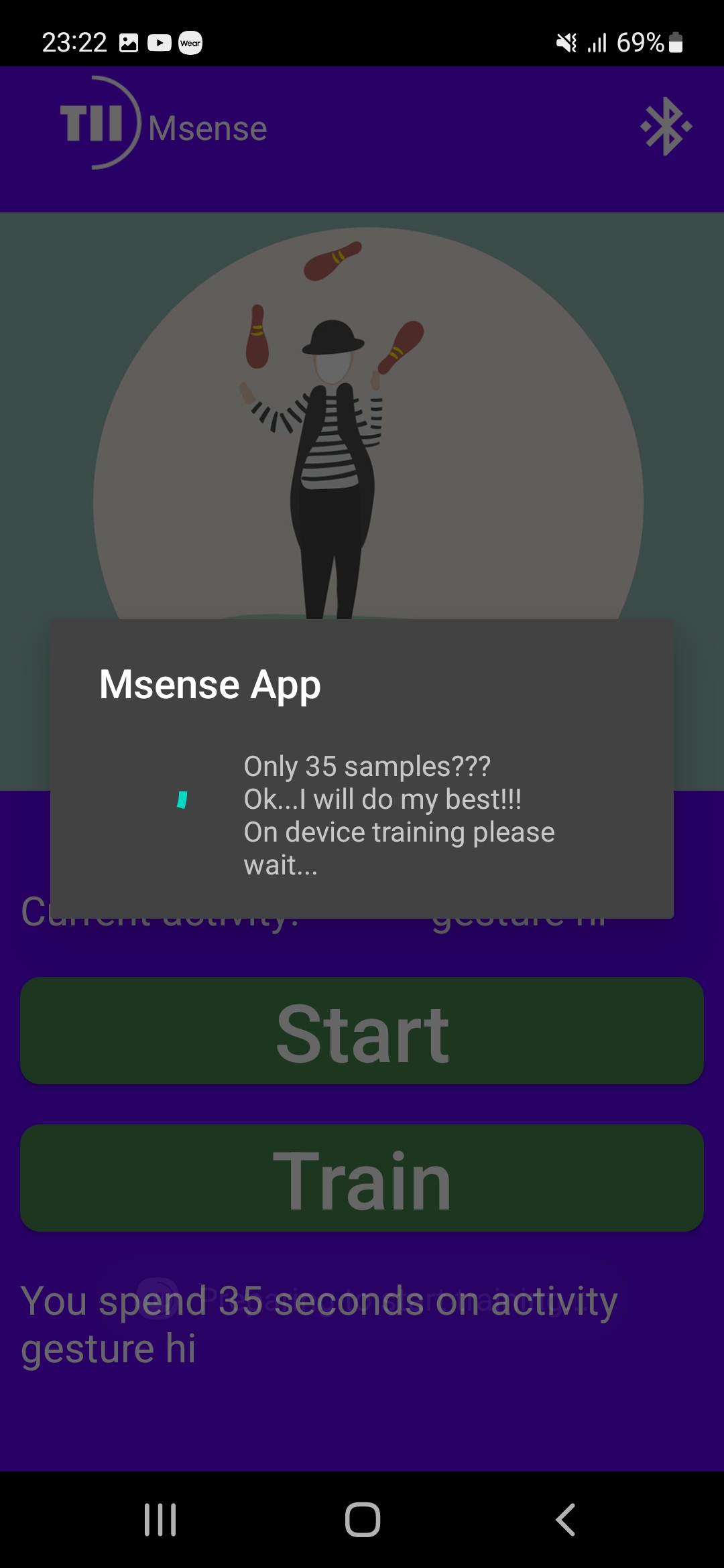}%
    \caption*{(d)}%
  \end{minipage}\hfil
  \begin{minipage}[t]{.175\linewidth}
    \includegraphics[width=\linewidth]{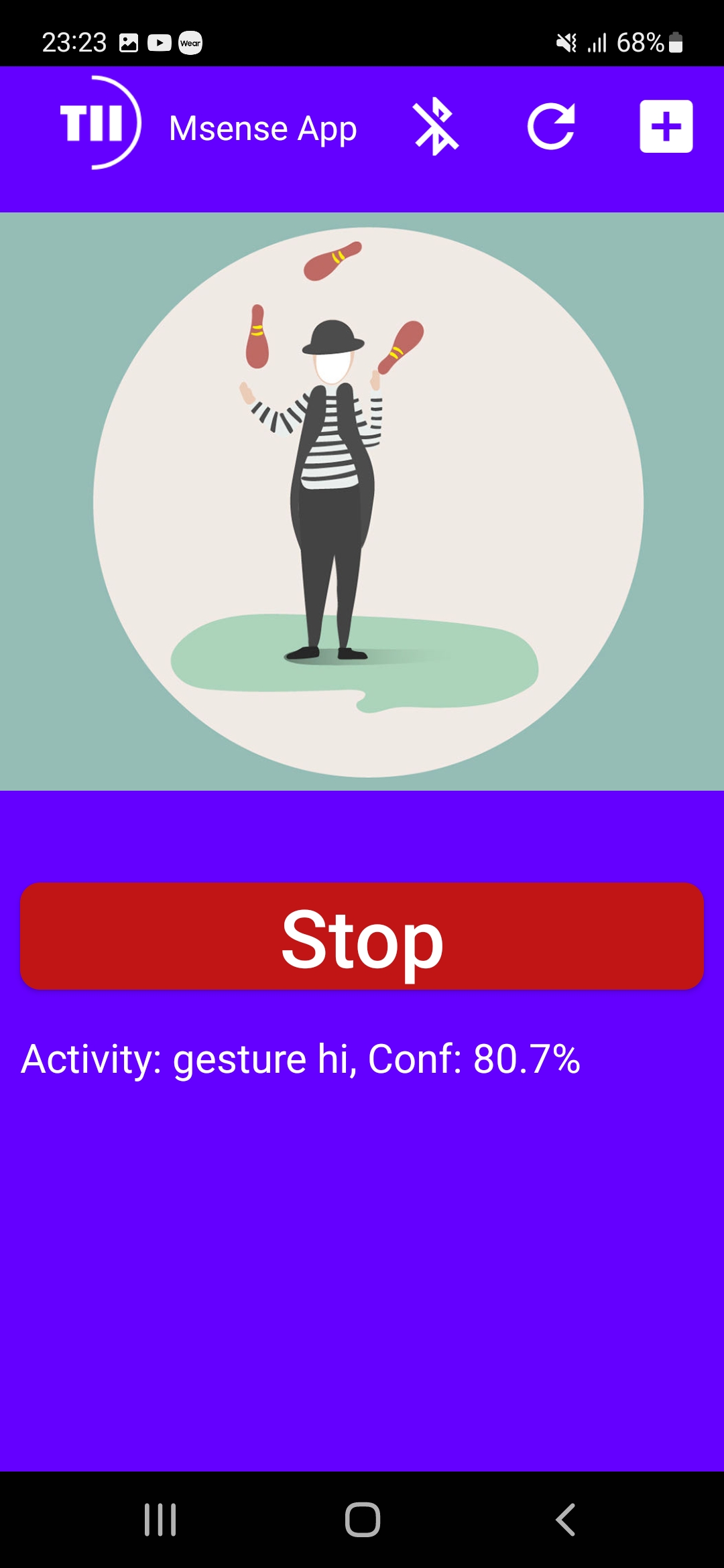}%
    \caption*{(e)}%
  \end{minipage}%
  \caption{GUI of the \textsf{MAGNETO} App: (a) Inference on \textit{Still} with the initial model, (b) Inference on \textit{Walk} with the initial model, (c) Collecting new activity data for \textit{Gesture Hi}, (d) Updating the Edge model, (e) Inference on new activity \textit{Gesture Hi}}
  \label{fig:activities}
\end{figure*}

%JZ: to describe what the audiences will experience during the demo. Add a paragraph "Demonstration Scenarios" for briefly describing the demo use cases (used data, user's action, what MAGNETO will show, etc.)
%JZ: Add a section/paragraph "Dataset statistics" for briefing the data collection & preprocessing 
During the demonstration of \textsf{MAGNETO} \footnote{The demo video can be found in \url{https://bit.ly/magneto_edbt2024}}, we focus on the following aspects: (i) Inference on the \textsf{Edge}, (ii) Incremental Learning of new activities on the fly, with no interaction with the Cloud. 
The demonstration device(s) will be disconnected from the Internet, ensuring the execution of the processes locally.
In order to improve the user's experience, the phone's outputs will be projected on a screen for a real-time visualization. 

\subsubsection{Real-time Inference}
As a first step, participants will be given a smartphone with \textsf{MAGNETO} installed, and try a few existing activities in real time, i.e., \textit{Drive}, \textit{E-scooter}, \textit{Run}, \textit{Still}, \textit{Walk}. 
%To demonstrate the \textsf{Edge} inference, the phone will be totally disconnected from the network. 
Figure~\ref{fig:activities}(a-b) show GUI examples for the real-time activity prediction. The participants will gain a clear understanding of the imperceptible prediction latency, which is only a few milliseconds.

\subsubsection{Incremental Learning}
With the same mobile phone, participants can collect new activity data with their personal behaviors, e.g., a greeting gesture as shown in Figure~\ref{fig:activities}(c) for recording a few seconds\footnote{A variety of activities can be used here. The listed ones are for illustration only}, and request \textsf{MAGNETO} to learn and integrate the activity into the existing model, as shown in Figure~\ref{fig:activities}(d). Participants can then check the new model's inference capability and check if it has managed to learn the new activity, see Figure~\ref{fig:activities}(e). 

We should remark that the entire data size that the demonstration needs on the \textsf{Edge} device (including support set, pre-processing, and the model) does not exceed $5 MB$, that is lower than two high definition pictures a smartphone can take, further motivating and highlighting the value of such a mechanism. 

%Additionally, for explainability purposes, we  provide some visualizations of the support set in the embedding space during the re-training, in order to visualise the training process the took place into the Edge device, see.

\section{Discussions and Conclusion}
\label{sec:conclusion}

There are multiple motivations for pushing the Human Activity Recognition (HAR) tasks to the \textsf{Edge}. As the sensing part, the Edge devices seamlessly integrate the data collection and processing tasks in the same device, providing extremely low latency. The isolated Edge environment naturally unblocks the strong privacy guarantees and personalization possibilities. However, \textsf{Edge} devices are extremely limited in terms of computational resources. This necessitates a careful design of machine learning models which can be efficient executed on the Edge devices. 

\textbf{Conclusion} In this paper, we present \textsf{MAGNETO}, an Edge AI platform designed for Human Activity Recognition (HAR) tasks, which provides rapid real-time response (low latency), enhanced privacy guarantees and personalization capability. Importantly, different from existing work or products in the market, \textsf{MAGNETO} allows adding new activities on the fly without the need for retraining the whole model. With a developed Android application, audiences can check the functionalities of the system in a user-friendly way. Even though \textsf{MAGNETO} is primarily designed for HAR tasks, its potential extends far beyond this application. Leveraging incremental learning, the system can adapt to diverse data types, such as time series, text, and voice. By adjusting its feature extractor or backbone model, the system offers a versatile solution adaptable to a wide range of scenarios.

%%
%% The acknowledgments section is defined using the "acks" environment
%% (and NOT an unnumbered section). This ensures the proper
%% identification of the section in the article metadata, and the
%% consistent spelling of the heading.
%\begin{acks}
%To Robert, for the bagels and explaining CMYK and color spaces.
%\end{acks}

%%
%% The next two lines define the bibliography style to be used, and
%% the bibliography file.
\bibliographystyle{ACM-Reference-Format}
\bibliography{references}

\end{document}